\documentclass[10pt,twocolumn,letterpaper]{article}

\usepackage{iccv}
\usepackage{times}
\usepackage{epsfig}
\usepackage{graphicx}
\usepackage{amsmath}
\usepackage{amssymb}
\usepackage{amsthm}
\usepackage{subcaption}
\usepackage{booktabs}
\usepackage{multirow}

\usepackage[breaklinks=true,bookmarks=false]{hyperref}

\iccvfinalcopy 

\def\iccvPaperID{4649} 

\ificcvfinal\fi

\begin{document}

\title{Domain Adaptation for Semantic Segmentation with Maximum Squares Loss}

\author{Minghao Chen, \, Hongyang Xue, \, Deng Cai\thanks{Corresponding author}\\ \\
	State Key Lab of CAD\&CG, College of Computer Science, Zhejiang University, Hangzhou, China\\
	Fabu Inc., Hangzhou, China\\
	Alibaba-Zhejiang University Joint Institute of Frontier Technologies, Hangzhou, China\\
	{\tt\small minghaochen01@gmail.com, hyxue@outlook.com, dengcai@cad.zju.edu.cn}
}

\maketitle
\ificcvfinal\thispagestyle{empty}\fi

\begin{abstract}
	Deep neural networks for semantic segmentation always require a large number of samples with pixel-level labels, which becomes the major difficulty in their real-world applications. To reduce the labeling cost, unsupervised domain adaptation (UDA) approaches are proposed to transfer knowledge from labeled synthesized datasets to unlabeled real-world datasets. Recently, some semi-supervised learning methods have been applied to UDA and achieved state-of-the-art performance. One of the most popular approaches in semi-supervised learning is the entropy minimization method. However, when applying the entropy minimization to UDA for semantic segmentation, the gradient of the entropy is biased towards samples that are easy to transfer. To balance the gradient of well-classified target samples, we propose the maximum squares loss. Our maximum squares loss prevents the training process being dominated by easy-to-transfer samples in the target domain. 
	Besides, we introduce the image-wise weighting ratio to alleviate the class imbalance in the unlabeled target domain. Both synthetic-to-real and cross-city adaptation experiments demonstrate the effectiveness of our proposed approach. The code is	released at \url{https://github.com/ZJULearning/MaxSquareLoss}.
\end{abstract}

\section{Introduction} \label{Section_1}

In the last few decades,  deep learning has achieved great success in the semantic segmentation task~\cite{deeplabv2,deeplabv3,deeplabv3+, FCN, PSPNet}. Researchers have made remarkable progress in promoting the performance of deep models on current datasets, such as PASCAL VOC-2012~\cite{PASCALVOC-2012} and Cityscapes~\cite{Cityscapes}. However, these real-world datasets with pixel-wise semantic labels demand an enormous amount of manual annotation work. For annotating Cityscapes, it takes $90$ minutes to label one image accurately~\cite{GTA5}. Because of this ``curse of dataset annotation'', real-world datasets for semantic segmentation often contain only a small number of samples, which inhibits the model's generalization to various real-world situations. One possible way to overcome this limitation is to utilize synthetic datasets, such as the Grand Theft Auto V (GTA5)~\cite{GTA5} and SYNTHIA~\cite{SYNTHIA}, which take much less time to label and own more samples containing various situations. However, the model trained on the synthetic dataset cannot generalize well to real-world examples via direct transfer, due to the large appearance gap between the two datasets.

\begin{figure}[t]
	\begin{center}
		\includegraphics[width=0.98\linewidth]{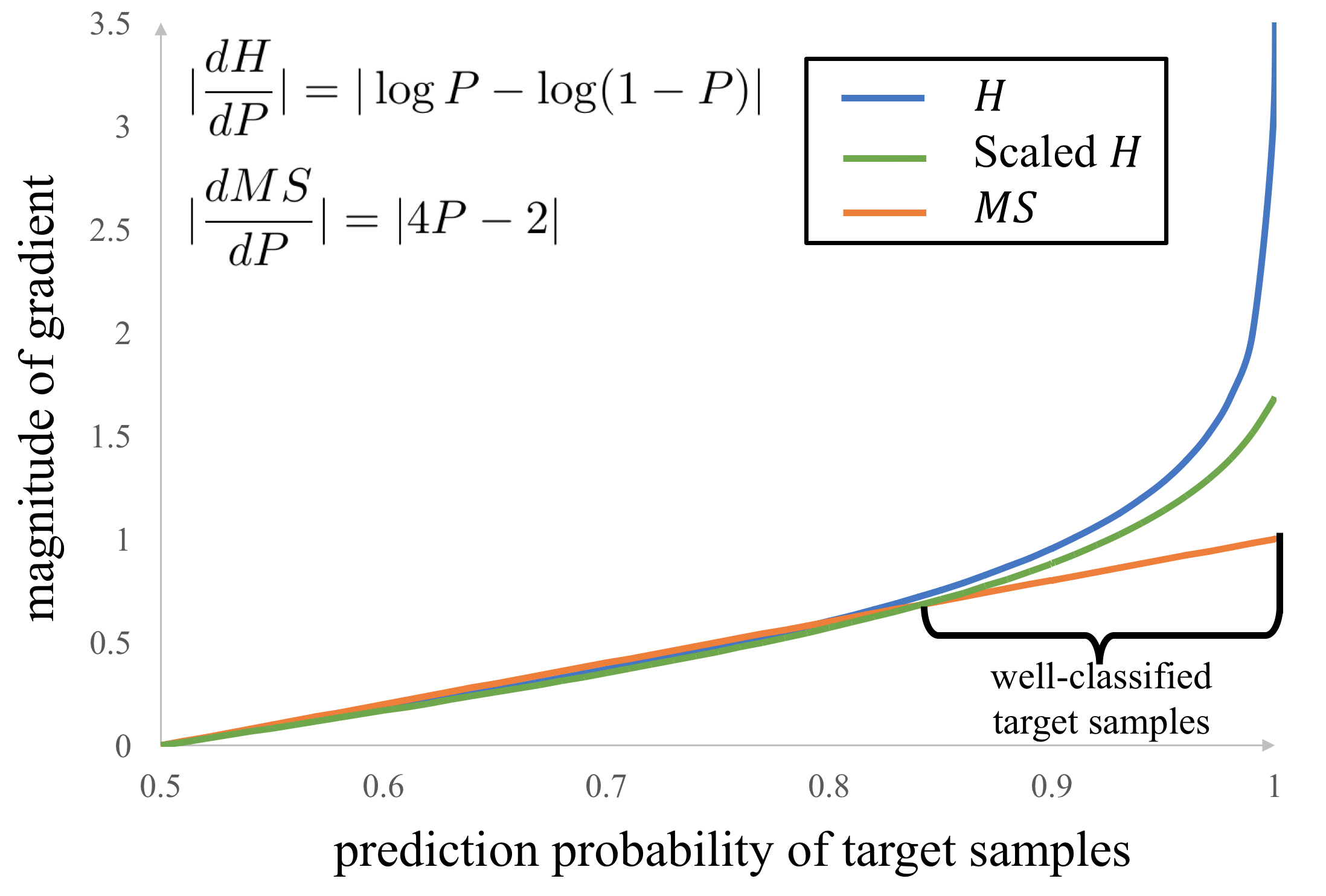}
	\end{center}
	\setlength{\abovecaptionskip}{-0.1cm}
	\setlength{\belowcaptionskip}{-0.2cm}
	\caption{In UDA, the gradient of the entropy minimization method ($H$) is focused on well-classified samples in the target domain. Consequently, we propose the maximum squares loss ($MS$), which is the negative sum of squared probabilities. The gradient of the maximum squares loss is linearly increasing, which reduces the gradient magnitude of samples that are easy to transfer and makes difficult samples be trained more efficiently.}
	\label{fig_001}
\end{figure}

Unsupervised domain adaptation (UDA) for semantic segmentation~\cite{CyCADA, AdaptSegNet, Self-training} is a task aiming at solving the above transfer problem. In UDA, the labeled synthetic dataset is known as the source domain, and the unlabeled real-world dataset is known as the target domain. The general idea of UDA is utilizing the unlabeled data from the target domain to help minimize the performance gap between these two domains. 

Recently, inspired by semi-supervised learning~\cite{Semi-supervisedLearningbyEntropyMinimization, Semi-supervisedPseudo-Label}, which also utilizes the unlabeled data, semi-supervised learning based UDA~\cite{self-ensembling, ADVENT, Self-training} approaches are introduced to align feature distributions between domains implicitly. These semi-supervised learning based approaches achieve state-of-the-art results in both classification~\cite{self-ensembling} and semantic segmentation~\cite{Self-training}. Entropy minimization~\cite{Semi-supervisedLearningbyEntropyMinimization}, which encourages unambiguous cluster assignments, is one of the most popular methods in semi-supervised learning. ADVENT~\cite{ADVENT} directly adopts the entropy minimization method to UDA for semantic segmentation, but their result is inferior to state-of-the-art approaches. 

By analyzing the gradient of the entropy minimization method, we find that higher prediction probability induces a larger gradient\footnotemark[1] for the target sample (Fig.~\ref{fig_001}). If we adopt the assumption in self-training~\cite{Self-training} that target samples with higher prediction probability are more accurate, areas with high accuracy will be trained more sufficiently than areas with low accuracy. Therefore, the entropy minimization method will allow for adequate training of samples that are easy to transfer, which hinders the training process of samples that are difficult to transfer.
This problem in the entropy minimization can be termed \textbf{probability imbalance}: classes that are easy to transfer have a higher probability, which results in a much larger gradient than classes that are difficult to transfer.
One simple solution is to replace the prediction probability $P$ in the entropy formula with $P_{scaled} = (1-2\gamma) P + \gamma$, in which $\gamma$ is the scale ratio (``Scaled $H$'' in Fig.~\ref{fig_001}). Then the maximum gradient can be bounded by the factor $\gamma$, instead of going to infinity. However, this method introduces an extra hyper-parameter $\gamma$, which is tricky to select.
\footnotetext[1]{In this paper, the gradient refers to the magnitude of the gradient.}

In this paper, we introduce a new loss, \textbf{the maximum squares loss}, to tackle the probability imbalance problem. Since the maximum squares loss has a linearly increasing gradient (Fig.~\ref{fig_001}), it can prevent high confident areas from producing excessive gradients. Meanwhile, we show optimizing our loss is equivalent to maximizing the Pearson $\chi^2$ divergence with the uniform distribution. Maximizing this divergence can achieve class-wise distribution alignment between source and target domains. 

Moreover, we notice the class imbalance in the unlabeled target domain. Due to unavailable labels in the target domain, we propose the image-wise weighting factor based on percentages of different classes in an image.
Last but not least, we utilize multi-level outputs to boost performance. We apply the idea in weakly-supervised learning~\cite{SelfGuided} to UDA and generate self-produced guidance to train the low-level feature.

The main contributions of this paper are as follows:
\begin{itemize}
	\item We discover the probability imbalance problem in the entropy minimization method of UDA, by analyzing the gradient of entropy. We propose the maximum squares loss with a linear growth gradient to balance the gradient of highly confident classes. 
	\item To tackle the class imbalance in the unlabeled target domain, we introduce the image-wise weighting factor, which is more suitable to UDA than conventional class weighting factors. 
	\item  Our approach can achieve competitive results with state-of-the-art methods under multiple UDA settings. It should be emphasized that our approach does not need additional structure or discriminator. Moreover, unlike self-training~\cite{Self-training}, our approach does not demand redundant computation to get pseudo-labels. 
\end{itemize}

\section{Related Work}
\textbf{Semantic Segmentation.} After years of research, semantic segmentation models based on deep neural networks (\eg, Deeplab~\cite{deeplabv2, deeplabv3, deeplabv3+}, PSPNet~\cite{PSPNet}) can achieve astonishing performance on the real-world datasets, \eg, PASCAL VOC-2012~\cite{PASCALVOC-2012}, and Cityscapes~\cite{Cityscapes}. Nevertheless, the performance heavily relies on high-quality labeled datasets, which need lots of manual effort. One possible way to reduce manual labeling cost is to adopt synthetic datasets constructed from the virtual world, \eg, SYNTHIA~\cite{SYNTHIA} and GTA5~\cite{GTA5}. However, due to the appearance difference between rendering and real images, there is a performance gap during the transfer from synthetic to real datasets. 

\textbf{Unsupervised Domain Adaptation.} Traditionally, unsupervised domain adaptation (UDA) \cite{DANN, DAN, JAN, ADDA, DDC, LearningSemanticRepresentationsforUDA} is studied to tackle the domain-shift problem between the labeled source domain and unlabeled target domain for the classification task. The core idea behind UDA is to minimize the divergence between the feature distributions of the source and target domains, which means to learn domain invariant features. The distribution divergence can be measured by Maximum Mean Discrepancy (MMD) based methods~\cite{DAN, JAN, DDC} or adversarial learning based methods~\cite{DANN, ADDA}. Apart from global distribution alignment, class-wise and conditional distribution alignments~\cite{JAN, LearningSemanticRepresentationsforUDA} are also widely studied.

\textbf{UDA for Semantic Segmentation.} For the semantic segmentation task, it is not suitable for direct adoption of approaches proposed for the classification task, due to the higher dimensional feature space. FCN in the wild~\cite{FCNsintheWild} firstly introduced the task of UDA for semantic segmentation, and tackled it with global feature alignment and label statistic matching. Output adaptation method~\cite{AdaptSegNet} adapted the structured output space to transfer the structured spatial knowledge. The conditional generator can be utilized to align the conditioned distribution~\cite{ConditionalGANDA}. Besides the adversarial methods, another idea is to transfer the style of real images to synthetic samples while keeping semantic labels. CyCADA~\cite{CyCADA} adopted CycleGAN~\cite{CycleGAN} to construct a labeled real-like dataset, which is more similar to the target dataset.

\textbf{Semi-supervised Learning Based Methods.} Recently, inspired by semi-supervised learning~\cite{Semi-supervisedLearningbyEntropyMinimization, Semi-supervisedPseudo-Label} which also utilizes the unlabeled data, there are several semi-supervised learning based methods~\cite{self-ensembling, self-ensemblingformedicalimagingsegmentation, Self-training, ADVENT} proposed for UDA task. Assuming that areas with higher prediction probability are more accurate, the class-balanced self-training~\cite{Self-training} generated pseudo labels based on class-wise thresholds. 

In semi-supervised learning study, it is concluded that the information content of unlabeled examples decreases as classes overlap~\cite{Semi-supervisedstudy2, Semi-supervisedstudy1}. Thus making unlabeled samples less ambiguous can help classes to be more separable, \eg, minimizing the conditional entropy~\cite{Semi-supervisedLearningbyEntropyMinimization}. ADVENT~\cite{ADVENT} adopted this idea in the UDA field and minimized the prediction entropy of the target sample.


\section{Methods}

In this section, we present our major contributions, \ie, the maximum squares loss, and the image-wise class-balanced weighting factor. In Section 3.1, we review UDA for semantic segmentation.  In Section 3.2, we illustrate the probability imbalance problem in the entropy minimization method for UDA and introduce our maximum squares loss. Then we reveal the benefit of maximum squares loss by the gradient analysis and explain the meaning of this loss from the perspective of $f$-divergence. Furthermore, in Section 3.3, we notice the class imbalance and solve it with our image-wise weighting factor. Last but not least, we apply the self-produced guidance to UDA, in Section 3.4.

\subsection{Overview of UDA}

In unsupervised domain adaptation (UDA), the labeled source domain is denoted as  $\mathcal{D}_S=\{(x_s, y_s) | x_s \in \mathbb { R } ^ { H \times W \times 3 }, y_s \in \mathbb { R } ^ { H \times W }\}$, and the unlabeled target domain is denoted as $\mathcal{D}_T=\{x_t | x_t \in \mathbb { R } ^ { H \times W \times 3 }\}$. The general objective function of UDA for semantic segmentation can be formulated as follows:
\begin{align}
\mathcal{L}(x_s, x_t) = \mathcal{L}_{CE}(p_s, y_s)+\lambda_{T}\mathcal{L}_{T}(x_t), \label{equ_001} \\
\mathcal{L}_{CE}(p_s, y_s)=-\frac{1}{N} \sum_{n=1}^{N} \sum_{c=1}^{C} y^{n,c}_s\log(p^{n,c}_s), \label{equ_002}
\end{align}
where $\mathcal{L}_{CE}$ is the cross entropy loss of source samples, $n$ represents a pixel point in the $H \times W$ space and $N=HW$ is the total number of pixels in a picture . $p^{n,c}_s$ is the model prediction probability of the class $c$ at point $n$ for sample $x_s$.  $\mathcal{L}_{T}(x_t)$ is the loss part for target samples.

\textbf{Entropy Minimization.} In the ~\cite{ADVENT}, they try to minimize the Shannon entropy of the target sample prediction. Thus, their objective function for target samples  is:
\begin{align}
\mathcal{L}_{T}(x_t) =- \frac{1}{N} \sum_{n=1}^{N} \sum_{c=1}^{C}  p^{n,c}_t \log(p^{n,c}_t). \label{equ_003}
\end{align}

For the sake of simplicity, we consider the binary classification case. Then the entropy formula and the gradient function of the entropy can be written as follows:
\begin{align}
H(p|x_t) &= - p\log p - (1-p)\log(1-p) , \label{equ_004} \\
|\frac{d H}{d p}| &=|\log p - \log(1-p)| . \label{equ_005}
\end{align}
After plotting the gradient function image on Fig.~\ref{fig_001}, we can see that the gradient of the high probability point is much larger than the mediate point. As a result, the key principle behind the entropy minimization method is that the training of target samples is guided by the high probability area, which is assumed to be more accurate.

\subsection{Maximum Squares Loss} \label{Section_3_2}
\begin{figure}[t]
	\begin{center}
		\includegraphics[width=0.98\linewidth]{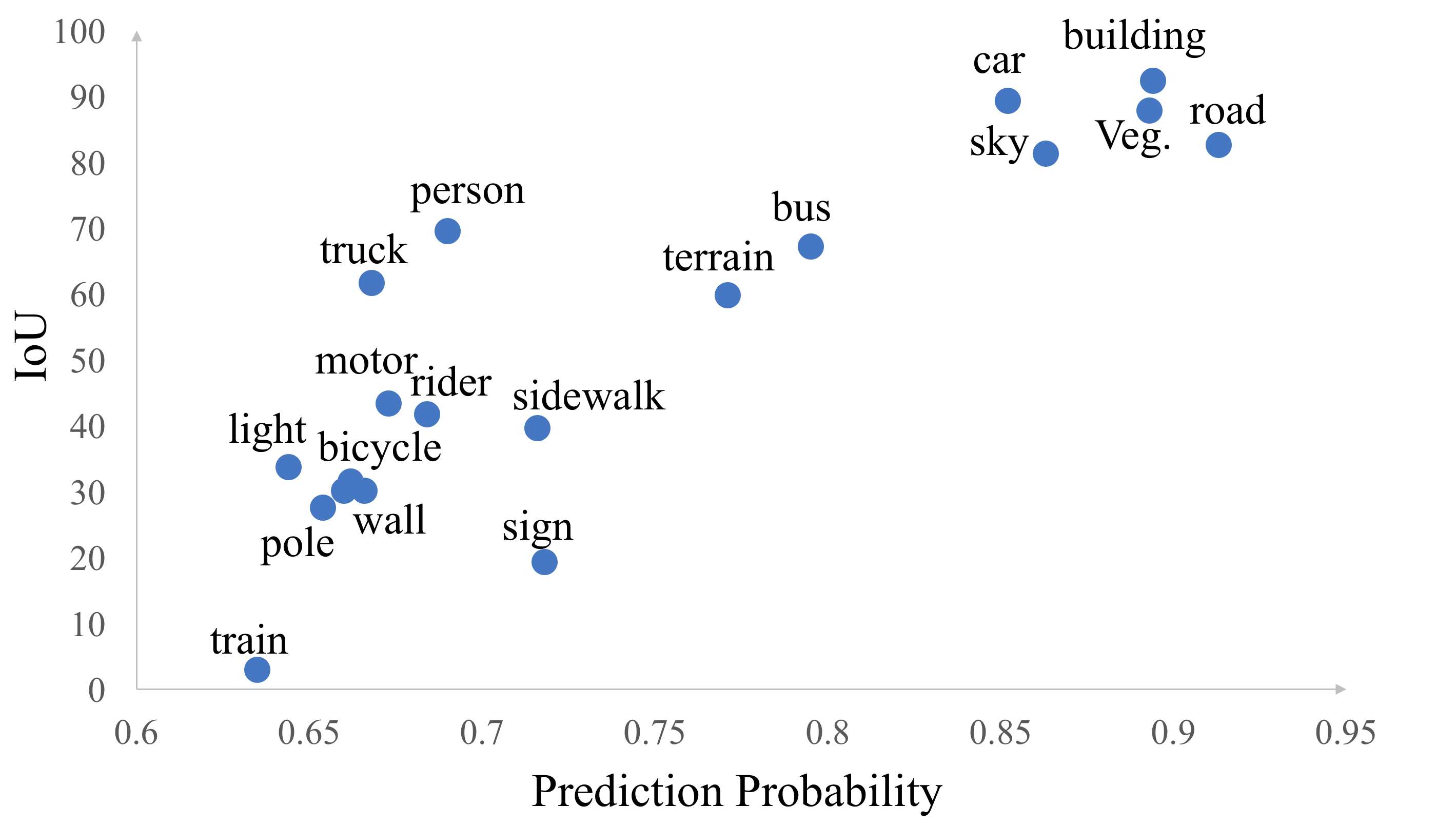}
	\end{center}
	\setlength{\abovecaptionskip}{-0.1cm}
	\setlength{\belowcaptionskip}{-0.1cm}
	\caption{From GTA5 to Cityscapes, the mean of prediction probability v.s. Intersection over Union(IoU) for each target class. They are almost linearly related. Thus well-classified classes (high IoU) have larger prediction probability.
	}
	\label{fig_002}
\end{figure}

\textbf{Probability Imbalance Problem.}
The probability of different classes varies widely. Classes with high accuracy always have higher prediction probabilities (Fig.~\ref{fig_002}). However, the gradient growth (Eq.~\ref{equ_005}) of the high probability point is approximated as $|\log p| (p \to 0)$, which will grow to infinity. Then the simple class will produce a much larger gradient on each pixel than the difficult class, resulting in the probability imbalance problem mentioned in Section~\ref{Section_1}. 
To remedy this problem, we define the \textbf{maximum squares loss} as:
\begin{align}
\mathcal{L}_{T}(x_t)=-\frac{1}{2N} \sum_{n=1}^{N} \sum_{c=1}^{C} ({p^{n,c}_t})^{2} . \label{equ_006}
\end{align}

\subsubsection{Benefit of Maximum Squares Loss} \label{Section_3_2_1}
For the binary classification case, we have the maximum squares loss and its gradient function as follows:
\begin{align}
MS(p|x_t) &= -p^2-(1-p)^2, \\
|\frac{d MS}{d p}| &=|4p-2| .
\end{align}
As the above equation shows, the gradient of the maximum square loss increases linearly (Fig.~\ref{fig_001}). It has a more balanced gradient for different classes than the entropy minimization method in the target domain. Areas with higher confidence still have larger gradients, but their dominant effects have been reduced, allowing other difficult classes to obtain training gradients. Therefore, equipped with the maximum square loss, we alleviate the probability imbalance in the entropy minimization. 

In the experiments (Section~\ref{section_4_3}), we show the maximum square loss does balance the training process of different samples and exceeds the entropy minimization method by a large margin.

\subsubsection{Interpretation from $f$-divergence View}

The target part loss $\mathcal{L}_{T}(x_t)$ can be treated as the distance between the model prediction distribution $p^{n,c}$ and uniform distribution: $\mathcal{U} = \frac{1}{C}$. Minimizing this distance will reduce the ambiguity of the target samples and help classes to be more separable~\cite{Semi-supervisedLearningbyEntropyMinimization}.

In probability theory, it is common to use $f$-divergence functions to measure the difference between distributions:
\begin{align}
D _ { f } ( p \| q ) = \sum_c q ( c ) f \left( \frac { p ( c ) } { q ( c ) } \right). \label{equ_007}
\end{align} 


\begin{figure}[t]
	\begin{center}
		\includegraphics[width=0.98\linewidth]{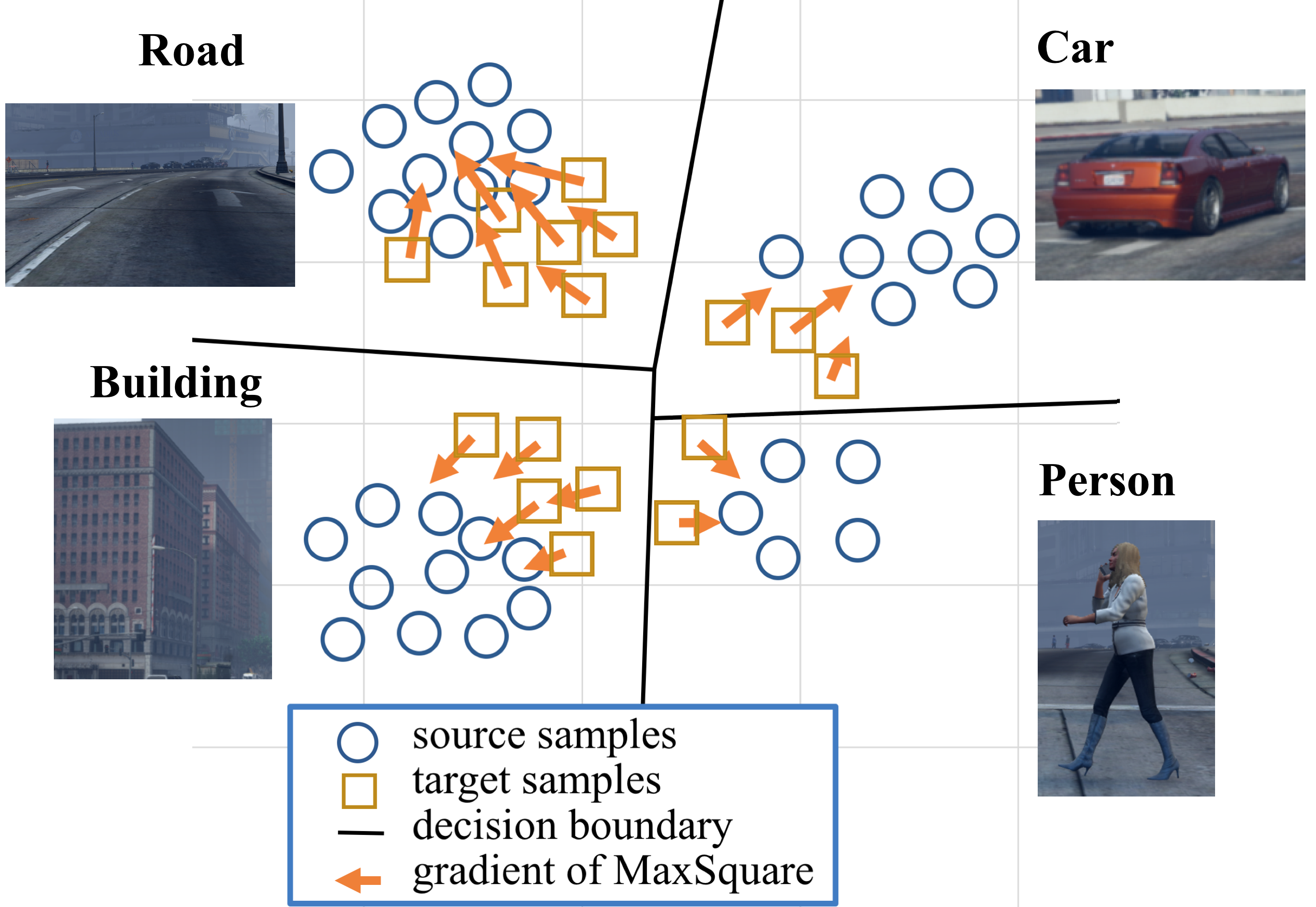}
	\end{center}
	\setlength{\abovecaptionskip}{-0.1cm}
	\setlength{\belowcaptionskip}{-0.0cm}
	\caption{The illustration of the effect of the maximum squares loss. Optimizing the maximum squares loss implicitly pushes the target sample features away from the decision boundary to the corresponding source feature distribution, which achieves class-wise distribution alignment.}
	\label{fig_003}
\end{figure}

We consider the Pearson $\chi^2$ divergence: $f(t) =t^2-1$ (or $f(t) =(t-1)^2$ equally). Then Eq.~\ref{equ_007} becomes:
\begin{align}
D _ { \chi^2 } ( {p^{n,c}} \| \mathcal{U} ) &= C \sum_c  ({p^{n,c}})^{2} -1. \label{equ_010}
\end{align}
Similar to entropy, the above equation is another metric for the ambiguity of the target sample. Maximize the Pearson $\chi^2$ divergence is equivalent to minimizing the objective function (Eq.~\ref{equ_006}). Maximizing the Pearson $\chi^2$ divergence with $\mathcal{U}$ will push the target features away from the decision boundary to the corresponding source feature distribution (Fig.~\ref{fig_003}). In this way, optimizing the maximum squares loss can achieve class-wise distribution alignment between two domains.

\subsection{Image-wise Class-balanced Weighting Factor} \label{Section_3_3}
\begin{figure}[t]
	\begin{center}
		\includegraphics[width=0.98\linewidth]{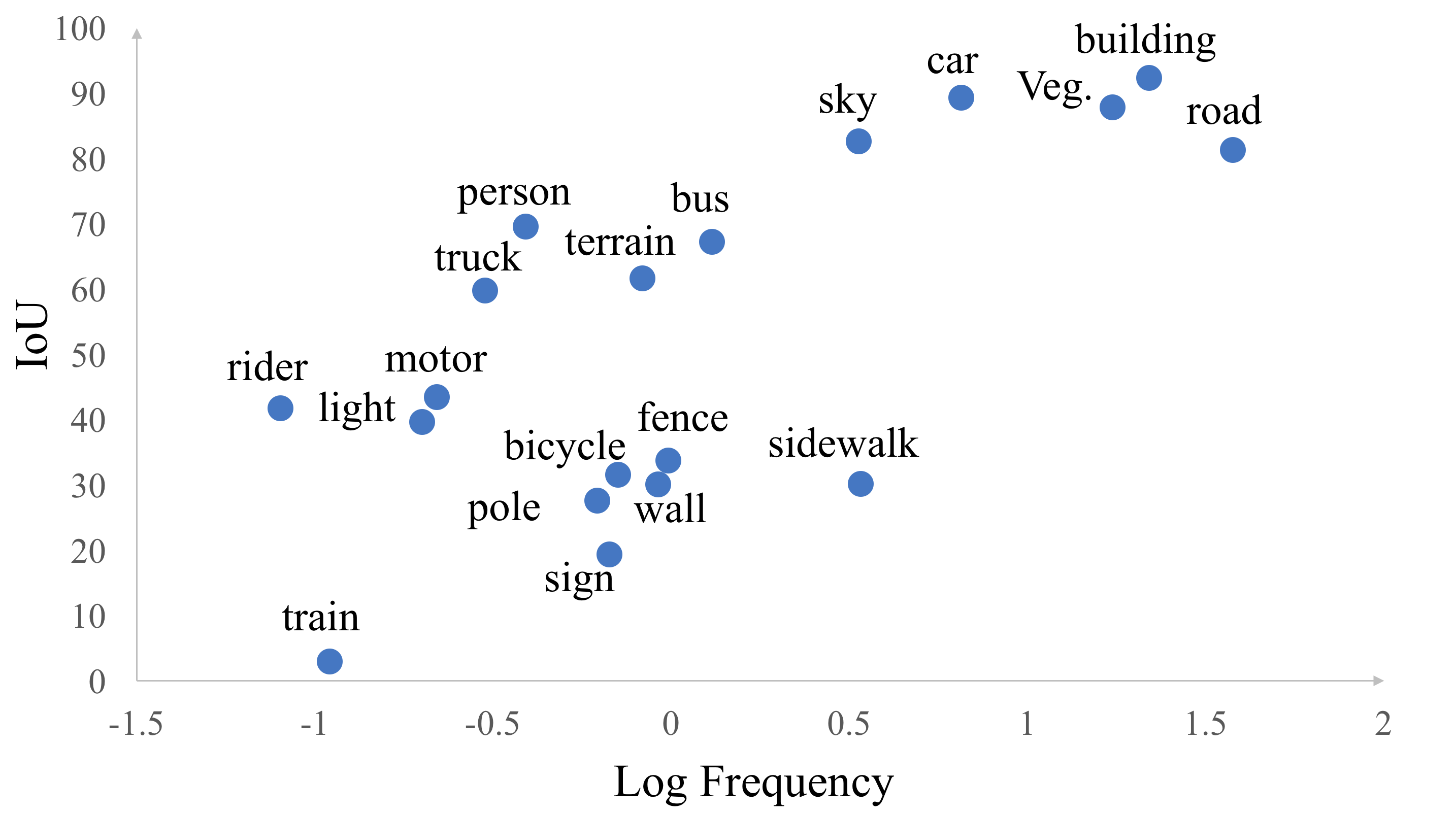}
	\end{center}
	\setlength{\abovecaptionskip}{-0.1cm}
	\setlength{\belowcaptionskip}{-0.1cm}
	\caption{From GTA5 to Cityscapes, the log frequency v.s. Intersection over Union (IoU) for each target class. They are almost linearly related. Thus well-classified classes (high IoU) have more pixels (high frequency). }
	\label{fig_004}
\end{figure}

As Fig.~\ref{fig_004} demonstrates, classes with higher accuracy always have more pixels on the label map, which leads to an imbalance in quantity. The regular method to balance the number of classes is to introduce weighting factor $\alpha_c$, which is usually set as the inverse class frequency~\cite{focalloss}. However, in the UDA task, there is no class label to calculate the class frequency. It is also not appropriate to replace the target class statistics with the class statistics on the source dataset, because there is no guarantee that the target domain will have the same class frequency as the source domain.

Instead of using the class frequency of the entire target dataset, we calculate them on each target image:
\begin{align}
{m}^{n,c^*} &=  \begin{cases} 
1  &\text { if } c^* = \underset { c } { \arg \max } p^{n,c} \\
0  &\text { otherwise }, \end{cases} \\
N^{c} &= \sum_n {m}^{n,c}.
\end{align}
In Eq.~\ref{equ_006}, we divide the sum by N to average the loss on the target image. Instead, we average the loss based on the number of classes $N^{c}$. Due to inaccurate predictions, 
interpolation between these two numbers is more stable:
\begin{align}
\mathcal{L}_{T}(x_t)=- \sum_{n=1}^{N} \sum_{c=1}^{C} \frac{1}{2{(N^{c})}^{\alpha}\times N^{(1-\alpha)}} {(p^{n,c}_t)}^2 , \label{equ_011}
\end{align}
where $\alpha$ is treated as a hyper-parameter to be selected by cross-validation.

\subsection{Multi-level Self-produced Guidance for UDA} \label{Section_3_4}

As mentioned in~\cite{AdaptSegNet}, adapting low-level feature can enhance the final performance. We extract the feature maps from the conv4 layer of ResNet~\cite{ResNet} and add an ASPP module to it as the low-level output. Then we extend the objective function of target samples as:
\begin{align}
\mathcal{L}_{T}(x_t) = \mathcal{L}_{T}^{final}(x_t) + \lambda_{low}\mathcal{L}_{T}^{low}(x_t),
\end{align}
where $\mathcal{L}_{T}^{final}(x_t)$ denotes the loss function of model final prediction for the target sample, \eg, the maximum squares loss (Eq.~\ref{equ_006}). Because the high-level output is more accurate than the low-level output, it is more reasonable to use the high-level output to guide the training of low-level features. As a result, we adopt the idea of the self-produced guidance learning~\cite{SelfGuided} in weakly-supervised learning. We first get the ensemble output $P_{ens}$ by averaging the output map of different levels, \ie, $P_{final}$ and $P_{low}$. Then we generate the self-produced guidance $\overline{y}^{n,c^*}_t$ by:
\begin{align}
\overline{y}^{n,c^*}_t =  \begin{cases} 
1  & \text { if } c^* = \underset { c } { \arg \max } \, p^{n,c}_{ens}, \\
& p^{n,c^*}_{final} > \delta \; or \; p^{n,c^*}_{low} > \delta  \\  
0  & \text { otherwise }, \end{cases} \label{equ_008}
\end{align}
where the choice of $\delta$ dose not effect the experimental result and we set $\delta=0.95$. We use this high-qualify guidance to guide the low-level training:
\begin{align}
\lambda_{low}\mathcal{L}_{T}^{low}(x_t) = \lambda_{low}\mathcal{L}_{CE}(p_{low}, \overline{y}^{n,c^*}_t).
\end{align}
In the experiment, we fix $\lambda_{low}=0.1$, the same as ~\cite{AdaptSegNet}.

\section{Experiment}

In this section, we first present the comparison between entropy minimization and maximum square loss on the classification task. Then, we conduct several experiments in the synthetic-to-real and cross-city settings to demonstrate the effectiveness of our approach in unsupervised domain adaptation for semantic segmentation. The code will be available at \url{https://github.com/ZJULearning/MaxSquareLoss}.

\begin{table*}[t]
	\centering
	\resizebox{0.65\textwidth}{11.0mm}{    
		\begin{tabular}{cccccccc}  
			\toprule
			Method & A $\to$ W & D $\to$ W &  W $\to$ D & A $\to$ D & D $\to$ A&  W $\to$ A & Avg    \\
			\midrule
			ResNet-50~\cite{ResNet} &  68.4$\pm$0.2 & 96.7$\pm$0.1  & 99.3$\pm$0.1  & 68.9$\pm$0.2 & 62.5$\pm$0.3 & 60.7$\pm$0.3 & 76.1    \\
			DANN~\cite{DANN}  &  82.0$\pm$0.4 & 96.9$\pm$0.2  & 99.1$\pm$0.1  & 79.7$\pm$0.4 & \textbf{68.2$\pm$0.4} & \textbf{67.4$\pm$0.5} & 82.2    \\
			EntMin            &  89.0$\pm$0.1 & 99.0$\pm$0.1  & \textbf{100.0$\pm$.0}  & 86.3$\pm$0.3 & 67.5$\pm$0.2 & 63.0$\pm$0.1 & 84.1    \\
			MaxSquare &  \textbf{92.4$\pm$0.5} & \textbf{99.1$\pm$0.1}  & \textbf{100.0$\pm$.0}  & \textbf{90.0$\pm$0.2} & 68.1$\pm$0.4 & 64.2$\pm$0.2 & \textbf{85.6}    \\
			\bottomrule
	\end{tabular}}
	\caption{Comparison between the entropy minimization and maximum square loss on Office-31.}
	\label{tab_004}
\end{table*}

\begin{table*}[t]
	\centering
	\resizebox{0.95\textwidth}{28mm}{    
		\setlength{\tabcolsep}{0.9mm}{
			\begin{tabular}{l| c | c c c c c c c c c c c c c c c c c c c | c }
				\toprule
				\multicolumn{22}{c}{GTA5$\to$Cityscapes} \\
				\midrule
				Method & Backbone & \rotatebox{90}{road} & \rotatebox{90}{sidewalk} & \rotatebox{90}{building} & \rotatebox{90}{wall} & \rotatebox{90}{fence} & \rotatebox{90}{pole} & \rotatebox{90}{light} & \rotatebox{90}{sign} & \rotatebox{90}{veg.} & \rotatebox{90}{terrain} & \rotatebox{90}{sky} & \rotatebox{90}{person} & \rotatebox{90}{rider} & \rotatebox{90}{car} & \rotatebox{90}{truck} & \rotatebox{90}{bus} & \rotatebox{90}{train} & \rotatebox{90}{motor} & \rotatebox{90}{bike} & mIoU (\%) \\
				\hline
				Source only~\cite{Self-training}
				&    Wider     & 70.0    & 23.7    & 67.8    & 15.4    & 18.1    & 40.2    & 41.9    & 25.3    & 78.8    & 11.7    & 31.4    & 62.9    & 29.8    & 60.1    & 21.5    & 26.8    & 7.7    & 28.1    & 12.0 & 35.4\\
				CBST~\cite{Self-training}
				& ResNet-38& 86.8 & 46.7 & 76.9 & 26.3 & 24.8 & 42.0 & 46.0 & 38.6 & 80.7 & 15.7 & 48.0 & 57.3 & 27.9 & 78.2 & 24.5 & 49.6 & 17.7 & 25.5 & 45.1 & 45.2\\
				CBST-SP~\cite{Self-training}
				& \cite{ResNet38} & 88.0    & \textbf{56.2}    & 77.0    & 27.4    & 22.4    & \textbf{40.7}    & \textbf{47.3}    & \textbf{40.9}    & 82.4    & 21.6    & 60.3    & 50.2    & 20.4    & 83.8    & 35.0    & \textbf{51.0}    & \textbf{15.2}    & 20.6    & 37.0 & 46.2 \\
				\hline
				
				AdaptSegNet~\cite{AdaptSegNet}
				& \multirow{3}{*}{ResNet101}     & 86.5 & 36.0 & 79.9 & 23.4 & 23.3 & 23.9 & 35.2 & 14.8 & 83.4 & 33.3    & 75.6 & 58.5 & 27.6 & 73.7 & 32.5 & 35.4 & 3.9 & 30.1 & 28.1    & 42.4 \\
				MinEnt~\cite{ADVENT}
				&     & 86.2 & 18.6 & 80.3 & 27.2 & 24.0 & 23.4 & 33.5 & 24.7 & 83.3 & 31.0 & 75.6 & 54.6 & 25.6 & \textbf{85.2} & 30.0 & 10.9 & 0.1 & 21.9 &  \textbf{37.1}    & 42.3    \\
				AdvEnt+MinEnt~\cite{ADVENT}        
				&     & 87.6 & 21.4 & 82.0 & \textbf{34.8} & \textbf{26.2} & 28.5 & 35.6 & 23.0 & 84.5 & 35.1 & 76.2 & 58.6 & \textbf{30.7} & 84.8 & 34.2 & 43.4 & 0.4 & 28.4 & 35.3    & 44.8    \\
				\hline
				Source only
				& \multirow{5}{*}{ResNet101}    & 71.4 & 15.3 & 74.0 & 21.1 & 14.4 & 22.8 & 33.9 & 18.6 & 80.7 & 20.9 & 68.5 & 56.6 & 27.1 & 67.4 & 32.8 & 5.6 & 7.7 & 28.4 & 33.8 & 36.9 \\
				MinEnt$^\dagger$
				& & 84.2 & 34.4 & 80.7 & 27.0 & 15.7 & 25.8 & 32.6 & 18.0 & 83.4 & 29.4 & 76.9 & 58.7 & 24.0 & 78.7 & 35.9 & 29.9 & 6.5 & 28.3 & 31.4 & 42.2  \\
				MaxSquare    
				& & 88.1 & 27.7 & 80.8 & 28.7 & 19.8 & 24.9 & 34.0 & 17.8 & 83.6 & 34.7 & 76.0 & 58.6 & 28.6 & 84.1 & 37.8 & 43.1 & 7.2 & 32.2 & 34.2 & 44.3 \\
				MaxSquare+IW    
				& & 89.3 & 40.5 & 81.2 & 29.0 & 20.4 & 25.6 & 34.4 & 19.0 & 83.6 & 34.4 & 76.5 & 59.2 & 27.4 & 83.8 & \textbf{38.4} & 43.6 & 7.1 & 32.2 & 32.5 & 45.2 \\
				MaxSquare+IW+Multi
				& & \textbf{89.4} & 43.0 & \textbf{82.1} & 30.5 & 21.3 &30.3 & 34.7 & 24.0 & \textbf{85.3} & \textbf{39.4} & \textbf{78.2} & \textbf{63.0} & 22.9 & 84.6 & 36.4 & 43.0 & 5.5 & \textbf{34.7} & 33.5 & \textbf{46.4} \\
				\bottomrule    
	\end{tabular}}}
	\caption{Results for GTA5-to-Cityscapes experiments. ``MaxSquare'' denotes our maximum squares loss method and ``MaxSquare+IW'' is the maximum squares loss combined with our image-wise weighting factor (Eq.~\ref{equ_011}). `` Multi'' denotes combining the multi-level self-guided method in Section~\ref{Section_3_4}. For comparison, we reproduce the result of entropy minimization method~\cite{ADVENT}, which is denoted as ``MinEnt$^\dagger$''. CBST~\cite{Self-training} adopts a wider ResNet model~\cite{ResNet38}, which is more powerful than the original ResNet~\cite{ResNet} that we adopt.}
	\label{tab_001}
\end{table*}

\subsection{Datasets} 
\textbf{Classification.}
\textit{Office-31}~\cite{office} is the most commonly used dataset for unsupervised domain adaptation, which contains 4,652 images and 13 categories collected from three domains: \textit{Amazon} (\textbf{A}), \textit{Webcam} (\textbf{W}) and \textit{DSLR} (\textbf{D}). We evaluate all methods across six domain adaptation tasks \textbf{A} $\to$ \textbf{W}, \textbf{D} $\to$ \textbf{W}, \textbf{W} $\to$ \textbf{D}, \textbf{A} $\to$ \textbf{D}, \textbf{D} $\to$ \textbf{A} and \textbf{W} $\to$ \textbf{A}.

\textbf{Semantic Segmentation.}
As for the transfer from synthetic datasets to real-world datasets, we consider Cityscapes~\cite{Cityscapes} as the target domain, and set GTA5~\cite{GTA5} or SYNTHIA~\cite{SYNTHIA} dataset as the source domain, which is same as the setting in previous works~\cite{AdaptSegNet, Self-training}. Cityscapes dataset contains 5,000 annotated images with $2048 \times 1024$ resolution taken from real urban street scenes. GTA5 dataset~\cite{GTA5} contains 24,966 annotated images with $1914 \times 1052$ resolution taken from the the GTA5 game. For SYNTHIA dataset, we use the SYNTHIA-RAND-CITYSCAPES subset consisting of 9,400 $1280 \times 760$ synthetic images.  During training, we use the labeled training sets of GTA5 or SYNTHIA as the source domain and the 2,975 images from Cityscapes training set without annotation as the target domain. We evaluate all methods on the 500 images from Cityscapes validation set.

In the evaluation, we adopt the Intersection-over-Union (IoU) of each class and the mean-Intersection-over-Union (mIoU) as performance metrics. We consider the IoU and mIoU of all 19 classes in the GTA5-to-Cityscapes case. While SYNTHIA only shares 16 classes with Cityscapes, we consider the IoU and mIoU of 16-class and 13-class in the SYNTHIA-to-Cityscapes case.

As for cross-city adaptation, we choose the training set of Cityscapes as the source domain and NTHU dataset~\cite{NTHU} as the target domain. The NTHU dataset consists of images with $2048 \times 1024$ resolution from four different cities: Rio, Rome, Tokyo, and Taipei. For each city, we use 3200 images without annotations as the target domain for training and 100 images labeled with 13 classes for evaluation. We consider the shared 13-class IoU and mIoU for evaluation.

\subsection{Implementation Details}

\textbf{Classification.}
We applied entropy minimization and maximum square loss to ResNet-50~\cite{ResNet}. We adopt the model pre-trained on ImageNet~\cite{ImageNet}, except the final classifier layer. We train the model using stochastic gradient descent (SGD) with momentum of 0.9. Following learning rate annealing strategy in ~\cite{DANN}, the learning rate is adjusted by $\eta_p=\frac{\eta_0}{(1+\alpha p)^{\beta}}$,  where p is the training progress linearly changing from 0 to 1, $\eta_0=0.01$, $\alpha=10$, $\beta=0.75$. We set the batch size to 128, half of which is source samples and half is target samples. We set $\lambda_{T}=0.3$ for maximum square loss and $\lambda_{T}=0.03$ for entropy minimization.

\textbf{Semantic Segmentation.}
As argued in~\cite{AdaptSegNet}, it is important to adopt a stronger baseline model to understand the effect of different adaption approaches and enhance the performance for the practical application. Therefore, in all experiment, we use Deeplabv2~\cite{deeplabv2} with ResNet-101~\cite{ResNet} backbones pre-trained on ImageNet~\cite{ImageNet} as our base model, which is the same as other works~\cite{AdaptSegNet, ADVENT}. 

Before the adaptation, we pre-train the network on the source domain for $70k$ steps to get a high-quality source trained network. We implement the algorithms using PyTorch~\cite{PyTorch} on a single NVIDIA $1080$Ti GPU. Due to memory limitations, we train the model with batch size $2$ (one from the source domain and one from the target domain).

Following~\cite{AdaptSegNet}, we train the model with Stochastic Gradient Descent (SGD) optimizer with learning rate $2.5 \times 10^{-4}$ , momentum $0.9$ and weight decay $5 \times10^{-4}$. We schedule the learning rate using ``poly'' policy: the learning rate is multiplied by $(1-\frac{iter}{max\_iter})^{0.9}$~\cite{deeplabv2}. We employ the random mirror and gaussian blur to augment data, the same as~\cite{PSPNet}. 

As for the selection of hyper-parameters, we set $\lambda_{T}=0.1$ in all experiments. In the experiments related to the image-wise weighting factor (Eq.~\ref{equ_011}), we fix $\alpha=0.2$.

\begin{figure}[t]
	\centering
	\setlength{\belowcaptionskip}{-0.4cm}
	\includegraphics[width=1\linewidth]{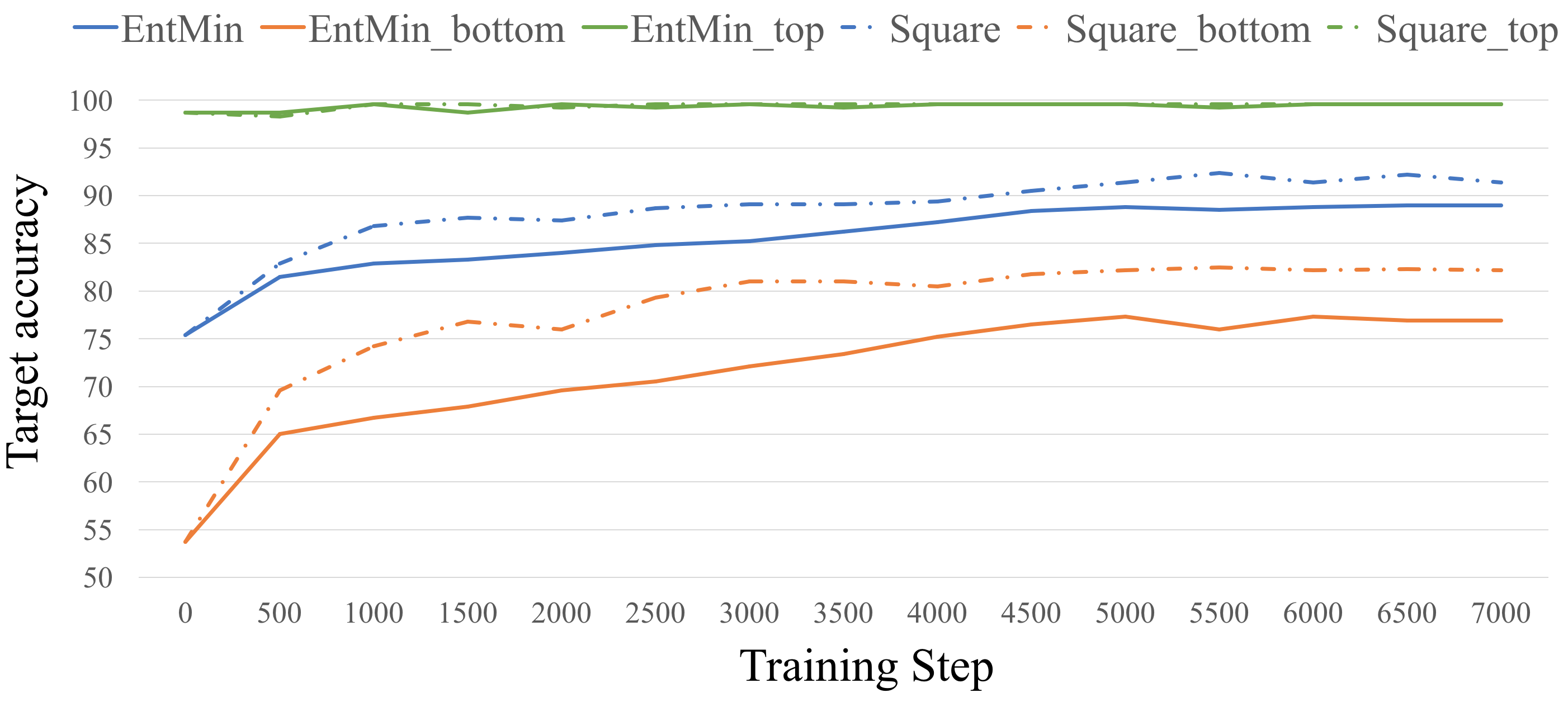}
	\caption{Accuracy of different difficulty samples on A$\to$W. For instance, ``EntMin\_bottom'' is the accuracy of the entropy minimization on the ``bottom set" (most difficult samples).}
	\label{fig_005}
\end{figure}

\subsection{Experiments on Classification}
\textbf{Results}
Tab.~\ref{fig_004} shows comparison results on \textit{office-31}. Although the results are uncompetitive with state-of-the-art methods, the maximum square loss (MaxSquare) exceeds the entropy minimization (EntMin) and DANN~\cite{DANN} by a large margin. Because the semantic segmentation task is much harder than the classification, this difference will be more apparent in the following semantic segmentation experiments.

\begin{table*}[h]
	\centering
	\resizebox{0.95\textwidth}{26.8mm}{    
		\setlength{\tabcolsep}{0.9mm}{
			\begin{tabular}{l| c | c c c c c c c c c c c c c c c c |c| c }
				\toprule
				\multicolumn{20}{c}{SYNTHIA$\to$Cityscapes} \\
				\midrule
				Method & Backbone & \rotatebox{90}{road} & \rotatebox{90}{sidewalk} & \rotatebox{90}{building} & \rotatebox{90}{wall*} & \rotatebox{90}{fence*} & \rotatebox{90}{pole*} & \rotatebox{90}{light} & \rotatebox{90}{sign} & \rotatebox{90}{veg.} & \rotatebox{90}{sky} & \rotatebox{90}{person} & \rotatebox{90}{rider} & \rotatebox{90}{car} & \rotatebox{90}{bus} & \rotatebox{90}{motor} & \rotatebox{90}{bike} & mIoU (\%) & mIoU* (\%) \\
				\hline
				Source only~\cite{Self-training}
				& Wider     & 32.6    & 21.5    & 46.5    & 4.8    & 0.1 & 26.5  & 14.8  & 13.1    & 70.8    & 60.3    & 56.6    & 3.5  & 74.1 & 20.4    & 8.9    & 13.1    & 29.2    & 33.6 \\
				CBST~\cite{Self-training}
				&  ResNet-38           & 53.6    & 23.7    & 75.0    & \textbf{12.5}    & 0.3    & \textbf{36.4}    & \textbf{23.5}    & \textbf{26.3}    & \textbf{84.8}    & 74.7    & \textbf{67.2}    & 17.5    & \textbf{84.5}    & 28.4    & 15.2    & \textbf{55.8}    & \textbf{42.5} & \textbf{48.4} \\
				\hline
				
				AdaptSegNet~\cite{AdaptSegNet}
				& \multirow{3}{*}{ResNet101} & 84.3 & \textbf{42.7} & 77.5  & -     & -         & -          & 4.7  & 7.0  & 77.9  & 82.5  & 54.3  & 21.0  & 72.3  & 32.2  & \textbf{18.9}  & 32.3  & -   & 46.7 \\
				MinEnt~\cite{ADVENT}        
				&                                         & 73.5 & 29.2 & 77.1 & 7.7 & 0.2 & 27.0 & 7.1 & 11.4 & 76.7 & 82.1 & 57.2 & 21.3    & 69.4 & 29.2 & 12.9 & 27.9 & 38.1 & 44.2    \\
				AdvEnt+MinEnt~\cite{ADVENT}        
				&                                         & \textbf{85.6}  & 42.2 & 79.7 & 8.7 & 0.4    & 25.9   & 5.4   & 8.1  & 80.4  & 84.1  & 57.9 & \textbf{23.8} & 73.3 & \textbf{36.4}    & 14.2 & 33.0  & 41.2 & 48.0    \\
				\hline
				Source only
				& \multirow{5}{*}{ResNet101} & 17.7 & 15.0 & 74.3 & 10.1 & 0.1 & 25.5 & 6.3 & 10.2 & 75.5 & 77.9 & 57.1 & 19.2 & 31.2 & 31.2 & 10.0 & 20.1 & 30.1 & 34.3\\
				MinEnt$^\dagger$
				&                                       & 67.8 & 28.3 & 79.0 & 4.8 & 0.1 & 24.7 & 4.0 & 7.3 & 81.7 & 84.1 & 58.9 & 19.4 & 75.9 & 36.2 & 10.4 & 26.1 & 38.0 & 44.5 \\
				MaxSquare    
				&                                       & 77.4 & 34.0 & 78.7 & 5.6 & 0.2 & 27.7 & 5.8 & 9.8 & 80.7 & 83.2 & 58.5 & 20.5 & 74.1 & 32.1 & 11.0 & 29.9 & 39.3 & 45.8 \\
				MaxSquare+IW    
				&                                        & 78.5 & 34.7 & 76.3 & 6.5 & 0.1 & 30.4 & 12.4 & 12.2 & 82.2 & \textbf{84.3} & 59.9 & 17.9 & 80.6 & 24.1 & 15.2 & 31.2 & 40.4 & 46.9  \\
				MaxSquare+IW+Multi
				&                                        & 82.9 & 40.7 & \textbf{80.3} & 10.2 & \textbf{0.8} & 25.8 & 12.8 & 18.2 & 82.5 & 82.2 & 53.1 & 18.0 & 79.0 & 31.4 & 10.4 & 35.6 & \textbf{41.4} & \textbf{48.2}  \\
				\bottomrule    
	\end{tabular}}}
	\setlength{\belowcaptionskip}{-0.2cm}
	\caption{Results for SYNTHIA-to-Cityscapes experiments.}
	\label{tab_002}
\end{table*}

\textbf{Verification of Maximum Square Loss.}
As shown in Section ~\ref{Section_3_2}, the maximum squares loss can make difficult samples be trained more efficiently than the entropy minimization. We use A$\to$W task to verify this conclusion experimentally. We first train the model on the source domain and mark the 30\% most confident samples in the test set as ``top set" and the 30\% least confident samples as ``bottom set". Then we fine-tune the model with EntMin or MaxSquare and record the accuracy on the test set, ``top set" and ``bottom set". As Fig.~\ref{fig_005} shows, there is no difference between the accuracy of two methods on the ``top set". However, the accuracy of MaxSquare on the ``bottom set" is much higher than EntMin. These results imply that the main improvement of MaxSquare to EntMin comes from the improvement of difficult samples. 

\subsection{GTA5 to Cityscapes} \label{section_4_3}

\subsubsection{Overall Results} \label{section_4_3_1}

Table~\ref{tab_001} summarizes the experimental results for GTA5-to-Cityscapes adaption comparing with state of the art methods~\cite{AdaptSegNet, ADVENT, Self-training}.  
As Table~\ref{tab_001} shows, equipped with ResNet-101 backbone, our ``MaxSquare+IW+Multi'' method achieves state-of-the-art performance. Compared with ``MaxSquare'', ``MaxSquare+IW'' 
shows better transfer results on small object classes, \eg, fence, person, truck, train, and motorbike. Besides, for those hard-to-transfer classes, \eg, terrain, bus and bike, ``MaxSquare'' performs better than the original entropy minimization method ``MinEnt$^\dagger$''~\cite{ADVENT}. However, we also find the ``MaxSquare' result for the well-classified road class is also improved than ``MinEnt$^\dagger$''. We explain this phenomenon that the maximum squares loss not only reduces gradients of easy-to-transfer classes but also reduces gradients of simple samples, which allows difficult samples from the road class to be trained more efficiently. This mechanism is similar to focal loss~\cite{focalloss}.

We notice that ``CBST-SP''~\cite{Self-training} achieves similar results to our approach. Their method assumes the spatial priors are shared between source and target domains. However, different datasets may have different spatial distributions, and their assumption does not always hold, which will be revealed in the experiment of cross-city adaptations.

\subsubsection{Analysis of Maximum Square Loss} \label{section_4_3_2}

We perform the following investigative experiments on GTA5 to Cityscapes. 

\begin{table}[t]
	\centering
	\resizebox{0.32\textwidth}{17.5mm}{    
		\begin{tabular}{cccc|c}  
			\toprule
			\multicolumn{5}{c}{GTA5$\to$Cityscapes} \\
			\midrule
			Entropy &  MaxSquare & IW & Multi    &  mIoU    \\
			\midrule
			\checkmark&                  &      &     & 42.2            \\
			~                &\checkmark &       &    & 44.3 \\
			\midrule
			~ \checkmark&                 &  \checkmark &  & 43.5 \\
			~                   &\checkmark &\checkmark& & 45.2 \\
			\midrule
			~                   &\checkmark     &   & \checkmark & 45.2 \\
			~                   &\checkmark     &\checkmark& \checkmark & \textbf{46.4} \\
			\bottomrule
	\end{tabular}}
	\caption{Ablation study.}
	\label{tab_005}
\end{table}

\begin{table}[t]
	\centering
	\resizebox{0.46\textwidth}{15.6mm}{    
		\begin{tabular}{ccccccc}  
			\toprule
			\multicolumn{7}{c}{GTA5$\to$Cityscapes} \\
			\midrule
			param $\lambda_T=$&  0.5  & 0.2 & 0.1  &  0.05 & 0.02   \\
			MaxSquare & 43.2 & 44.1 &  \textbf{44.3} & 43.7 & 43.0 &\\
			\midrule
			param $\alpha=$ & 0 & 0.1 & 0.15 & 0.2 & 0.25 & 0.3 \\
			MaxSquare+IW & 44.3 & 44.8 & \textbf{45.2} & \textbf{45.2} & 44.8 & 44.4 \\
			\midrule
			param $\delta=$ & 0.98 &  0.95 &  0.9 & 0.8    \\
			MaxSquare+IW+Multi &  \textbf{46.4} & \textbf{46.4}  & 46.2 & 46.1 & &  \\
			\bottomrule
	\end{tabular}}
	\setlength{\belowcaptionskip}{-0.4cm}
	\caption{Parameter sensitivity analysis.}
	\label{tab_006}
\end{table}


\begin{table*}[t]
	\centering
	\resizebox{0.8\textwidth}{54mm}{    
		\setlength{\tabcolsep}{0.9mm}{
			\begin{tabular}{l| c | c c c c c c c c c c c c c | c c }
				\toprule
				\multicolumn{16}{c}{ Cross-City Adaptation } \\
				\midrule
				City & Method & \rotatebox{90}{road} & \rotatebox{90}{sidewalk} & \rotatebox{90}{building} & \rotatebox{90}{light} & \rotatebox{90}{sign} & \rotatebox{90}{veg.} & \rotatebox{90}{sky} & \rotatebox{90}{person} & \rotatebox{90}{rider} & \rotatebox{90}{car} & \rotatebox{90}{bus} & \rotatebox{90}{motor} & \rotatebox{90}{bike} & mIoU (\%) \\
				\hline
				\multirow{6}{*}{Rome} 
				
				&     Cross city~\cite{NTHU}    & 79.5 & 29.3 & 84.5 & 0.0 & 22.2 & 80.6 & 82.8 & 29.5 & 13.0 & 71.7 & 37.5 & 25.9 & 1.0 & 42.9 \\
				
				&     CBST~\cite{Self-training}    & \textbf{87.1} & \textbf{43.9} & \textbf{89.7} & 14.8 & \textbf{47.7} & 85.4 & 90.3 & 45.4 & \textbf{26.6} & \textbf{85.4} & 20.5 & 49.8 & \textbf{10.3} & 53.6 \\
				
				&     AdaptSegNet~\cite{AdaptSegNet}    & 83.9 & 34.2 & 88.3 & 18.8 & 40.2 & \textbf{86.2} & \textbf{93.1} & \textbf{47.8} & 21.7 & 80.9 & 47.8 & 48.3 & 8.6 & 53.8\\
				
				& Source only  & 85.0 & 34.7 & 86.4 & 17.5 & 39.0 & 84.9 & 85.4 & 43.8 & 15.5 & 81.8 & 46.3 & 38.4 & 4.8 & 51.0 \\
				
				&      MaxSquare     & 80.0 & 27.6 & 87.0 & \textbf{20.8} & 42.5 & 85.1 & 92.4 & 46.7 & 22.9 & 82.1 & 53.5 & \textbf{50.8} & 8.8 & 53.9 \\
				
				&      MaxSquare+IW & 82.9 & 32.6 & 86.7 & 20.7 & 41.6 & 85.0 & 93.0 & 47.2 & 22.5 & 82.2 & \textbf{53.8} & 50.5 & 9.9 & \textbf{54.5} \\
				\hline
				\multirow{6}{*}{Rio} 
				
				&     Cross city~\cite{NTHU}    & 74.2 & 43.9 & 79.0 & 2.4 & 7.5 & 77.8 & 69.5 & 39.3 & 10.3 & 67.9 &  \textbf{41.2} & 27.9 & 10.9 & 42.5 \\
				
				&     CBST~\cite{Self-training}    &  \textbf{84.3} &  \textbf{55.2} & 85.4 &  \textbf{19.6} &  \textbf{30.1} & 80.5 & 77.9 & 55.2 & 28.6 &  \textbf{79.7} & 33.2 & 37.6 & 11.5 & 52.2 \\
				
				&     AdaptSegNet~\cite{AdaptSegNet}    & 76.2 & 44.7 & 84.6 & 9.3 & 25.5 & 81.8 & 87.3 &  \textbf{55.3} & 32.7 & 74.3 & 28.9 &  \textbf{43.0} & 27.6 & 51.6\\
				
				& Source only  & 74.2 & 42.2 & 84.0 & 12.1 & 20.4 & 78.3 & 87.9 & 50.1 & 25.6 & 76.6 & 40.0 & 27.6 & 17.0 & 48.9 \\
				
				&      MaxSquare     & 70.9 & 39.2 &  \textbf{85.6} & 14.5 & 19.7 &  \textbf{81.8} & \textbf{88.1} & 55.2 & 31.5 & 77.2 & 39.3 & 43.1 &  \textbf{30.1} & 52.0 \\
				
				&      MaxSquare+IW& 76.9 & 48.8 & 85.2 & 13.8 & 18.9 & 81.7 &  \textbf{88.1} & 54.9 &  \textbf{34.0} & 76.8 & 39.8 & 44.1 & 29.7 &  \textbf{53.3} \\
				\hline
				\multirow{6}{*}{Tokyo} 
				
				&     Cross city~\cite{NTHU}    & 83.4 & \textbf{35.4} & 72.8 & 12.3 & 12.7 & 77.4 & 64.3 & 42.7 & 21.5 & 64.1 & \textbf{20.8} & 8.9 & 40.3 & 42.8 \\
				
				&     CBST~\cite{Self-training}    & \textbf{85.2} & 33.6 & \textbf{80.4} & 8.3 & \textbf{31.1} & \textbf{83.9} & 78.2 & 53.2 & 28.9 & \textbf{72.7} & 4.4 & 27.0 & 47.0 & 48.8 \\
				
				&     AdaptSegNet~\cite{AdaptSegNet}    & 81.5 & 26.0 & 77.8 & \textbf{17.8} & 26.8 & 82.7 & \textbf{90.9} & 55.8 & \textbf{38.0} & 72.1 & 4.2 & 24.5 & \textbf{50.8} & 49.9\\
				
				& Source only  & 81.4 & 28.4 & 78.1 & 14.5 & 19.6 & 81.4 & 86.5 & 51.9 & 22.0 & 70.4 & 18.2 & 22.3 & 46.4 & 47.8 \\
				
				&      MaxSquare     & 79.3 & 28.5 & 78.3 & 14.5 & 27.9 & 82.8 & 89.6 & 57.3 & 31.9 & 71.9 & 6.0 & 29.1 & 49.2 & 49.7 \\
				
				&      MaxSquare+IW& 81.2 & 30.1 & 77.0 & 12.3 & 27.3 & 82.8 & 89.5 & \textbf{58.2} & 32.7 & 71.5 & 5.5 & \textbf{37.4} & 48.9 & \textbf{50.5} \\
				\hline
				\multirow{6}{*}{Taipei} 
				
				&     Cross city~\cite{NTHU}    & 78.6 & 28.6 & 80.0 & 13.1 & 7.6 & 68.2 & 82.1 & 16.8 & 9.4 & 60.4 & 34.0 & 26.5 & 9.9 & 39.6 \\
				
				&     CBST~\cite{Self-training}    & \textbf{86.1} & \textbf{35.2} & 84.2 & 15.0 & \textbf{22.2} & 75.6 & 74.9 & 22.7 & \textbf{33.1} & \textbf{78.0} & 37.6 & \textbf{58.0} & 30.9 & 50.3 \\
				
				&     AdaptSegNet~\cite{AdaptSegNet}    & 81.7 & 29.5 & 85.2 & 26.4 & 15.6 & 76.7 & 91.7 & 31.0 & 12.5 & 71.5 & 41.1 & 47.3 & 27.7 & 49.1\\
				
				& Source only  & 82.6 & 33.0 & \textbf{86.3} & 16.0 & 16.5 & 78.3 & 83.3 & 26.5 & 8.4 & 70.7 & 36.1 & 47.9 & 15.7 & 46.3 \\
				
				&      MaxSquare    & 81.2 & 32.8 & 85.4 & \textbf{31.9} & 14.7 & \textbf{78.3} & \textbf{92.7} & 28.3 & 8.6 & 68.2 & 42.2 & 51.3 & 32.4 & 49.8 \\
				
				&      MaxSquare+IW& 80.7 & 32.5 & 85.5 & 32.7 & 15.1 & 78.1 & 91.3 & \textbf{32.9} & 7.6 & 69.5 & \textbf{44.8} & 52.4 & \textbf{34.9} & \textbf{50.6} \\
				\hline
				\bottomrule    
	\end{tabular}}}
	\setlength{\abovecaptionskip}{0.3cm}
	\setlength{\belowcaptionskip}{-0.3cm}
	\caption{Results for Cross-City experiments.}
	\label{tab_003}
\end{table*}

\textbf{Ablation Study.}
We investigate the effect of the image-wise weighting factor introduced in Section~\ref{Section_3_3}. When combined with the image-wise weighting factor (IW), performances of the entropy minimization and the maximum squares are improved by nearly 1 point (Tab.~\ref{tab_005}). As a result, the image-wise weighting factor is a robust solution to the class imbalance in the unlabeled target domain.

We also study the effect of the multi-level self-produced guidance in Section~\ref{Section_3_4}. As Table~\ref{tab_005} demonstrates, utilizing multi-level output can significantly improve the final performance.

\textbf{Parameter Sensitivity Analysis.} 
We show the sensitivity analysis of parameters $\lambda_{T}$, $\alpha$ and $\delta$ in Tab~\ref{tab_006}. Too large or too small $\lambda_{T}$ cannot take advantage of the maximum square loss. We empirically choose $\lambda_{T}=0.1$. As the table shows, ``MaxSquare+IW'' with different $\alpha$ always yields better performance than ``MaxSquare'', which shows that the image-wise weighting factor is robust to the hyper-parameter $\alpha$. Meanwhile, the choice of $\delta$ does not affect the result significantly, as mentioned in~\ref{Section_3_4}.

\subsection{SYNTHIA to Cityscapes} 

Following the evaluation protocol of other works~\cite{ADVENT, Self-training}, we evaluate the IoU and mIoU of the shared 16 classes between two datasets and the 13 classes excluding the classes with $^*$. As Table~\ref{tab_002} shows, our methods achieve competitive results to other methods. ``MaxSquare+IW'' surpasses ``MaxSquare'' method on the several small object classes, \eg, traffic light, traffic sign, and motorbike.

\subsection{Cross City Adaptation} 

To show the efficiency of our methods for smaller domain shift, we conduct our experiment on the NTHU dataset with ResNet-101 backbone. We consider the IoU and mIoU of shared 13 classes for evaluation. Table~\ref{tab_003} shows the results of transferring from Cityscapes to the four cities in the NTHU dataset. In all four adaptation experiments, our ``MaxSquare+IW'' outperforms the other most advanced methods by about 1 point. These excellent results demonstrate the effectiveness of our maximum squares loss and our image-wise weighting factor. Moreover, unlike self-training~\cite{Self-training}, our approach does not assume that source and target domains share the same spatial priors. Therefore, our method is robust to various transfer settings.

\section{Conclusion}
In this paper, we demonstrate the probability imbalance problem when applying the entropy minimization method to UDA for semantic segmentation. We propose the maximum squares loss to prevent easy-to-transfer classes from dominating the training on the target domain. We show that optimizing the maximum squares loss is equivalent to maximizing the Pearson $\chi^2$ divergence with the normal distribution. As for the class imbalance in the target domain, we propose to compute class weighting factor for each image, based on the prediction quantity of each class. The synthetic-to-real and cross-city adaption experiments show that our method can achieve state-of-the-art performance, without the discriminator in adversarial learning methods.

\subsection*{Acknowledgments}

This work was supported in part by the National Nature Science Foundation of China (Grant Nos: 61751307) and the National Youth Top-notch Talent Support Program.

{\small
	\bibliographystyle{ieee_fullname}
	\bibliography{egbib}
}

\end{document}